\documentclass{article}
\usepackage{ijcai19}

\usepackage{times}
\usepackage{soul}
\usepackage{url}
\usepackage[hidelinks]{hyperref}
\usepackage[utf8]{inputenc}
\usepackage[small]{caption}
\usepackage{graphicx}
\usepackage{amsmath}
\usepackage{booktabs}
\usepackage{algorithm}
\usepackage{algorithmic}
\usepackage{makecell}
\usepackage{graphicx}
\usepackage{url}
\usepackage{multirow}

\title{Learning to Select Knowledge for Response Generation in Dialog Systems}

\author{
Rongzhong Lian$^\dag$, Min Xie$^\ddag$, Fan Wang$^\dag$, Jinhua Peng$^\dag$, Hua Wu$^\dag$
\affiliations
  $^\dag$ Baidu Inc. China \\
  $^\ddag$The Hong Kong University of Science and Technology
\emails
  {\{lianrongzhong, wangfan04, pengjinhua, wu\_hua\}@baidu.com} \\
  mxieaa@cse.ust.hk
  }

\begin{document}
\maketitle
\begin{abstract}
  End-to-end neural models for intelligent dialogue systems suffer from the problem of generating uninformative responses.
  Various methods were proposed to generate more informative responses by leveraging external knowledge. 
  However, 
  few previous work has focused on selecting appropriate knowledge in the learning process.
  The inappropriate selection of knowledge could prohibit the model from learning to make full
 use of the knowledge.
  Motivated by this, 
  we propose an end-to-end neural model which employs a novel knowledge selection mechanism
  where both \emph{prior and posterior distributions over knowledge} are used to facilitate knowledge selection.
  %
  Specifically, a posterior distribution over knowledge is inferred from both utterances and responses, and 
  it ensures the appropriate selection of knowledge during the training process. 
  Meanwhile, a prior distribution, which is inferred from utterances only,  is used to
  approximate the posterior distribution 
  so that appropriate knowledge can be selected even without responses during the inference process.
  Compared with the previous work, 
  our model can better incorporate appropriate knowledge in response generation.
  %
  %
  Experiments on both automatic and human evaluation verify
  the superiority of our model over previous baselines.
\end{abstract}


\section{Introduction}
End-to-end neural generative models attract much attention as a potential solution to open-domain dialogue systems.
The sequence-to-sequence (Seq2Seq) model \cite{shang2015neural,vinyals2015neural,cho2014learning} has achieved success in generating fluent responses. 
However, it tends to produce less informative responses, such as \textit{``I don't know''} and \textit{``That's cool''}, resulting in less attractive conversations. 

Variety of improvements \cite{zhou2018commonsense,ghazvininejad2018knowledge,liu2018knowledge} have been proposed toward informative dialogue generation, by leveraging external knowledge, including unstructured texts or structured data such as knowledge graphs.  
For example, the commonsense model proposed in \cite{zhou2018commonsense} took commonsense knowledge into account, which is served as knowledge background to facilitate conversation understanding.
The recently created datasets
Persona-chat \cite{zhang2018personalizing} and Wizard-of-Wikipedia \cite{dinan2018wizard} introduced conversation-related knowledge (e.g., the personal profiles in Persona-chat) in response generation 
where knowledge is used to direct conversation flow. 
Dinan et al.~\shortcite{dinan2018wizard} used ground-truth knowledge to guide knowledge selection, which demonstrates improvements over those not using such information.
However, ground-truth knowledge is difficult to obtain in reality.

Most of existing researches focused on selecting knowledge based on the semantic similarity (e.g., graph attention \cite{zhou2018commonsense}) between input utterances and knowledge.
This kind of semantic similarity is regarded as a \emph{prior distribution over knowledge}. 
However, a prior distribution cannot effectively guide appropriate knowledge selection since different knowledge can be used to generate diverse responses for the same input utterance. 
In contrast, given a specific utterance and response pair, 
the \emph{posterior distribution over knowledge}, which is inferred from both the utterance and the response (instead of the utterance only), can provide effective guidance on knowledge selection 
since the actual knowledge used in the response is considered. 
The discrepancy between the prior and posterior distributions brings difficulties in the learning process: 
the model could hardly select appropriate knowledge simply based on the prior distribution and without response information, 
it is difficult to obtain the correct posterior distribution during the inference process. 
This kind of discrepancy would stop the model from learning to generating proper responses by utilizing appropriate knowledge.



\begin{table}[bt]
\centering
\small
\begin{tabular}{c|l} \hline
      Utterance & \makecell[l]{Hi! I do not have a favorite band but my\\ favorite reading is twilight. } \\ \hline \hline
      \makecell{Profiles/\\Knowledge}  & 
      \makecell[l]{K1. I love the band red hot chili peppers. \\ \hline
      K2. My feet are size six women s.\\ \hline
      K3. I want to be a journalist but instead I \\ sell washers at sears.
      } \\ \hline \hline
      \makecell{R1 (no knowledge)} & What do you do for a living? \\ \hline
      \makecell{R2 (use K2)} & I bought a pair of shoes of size six women. \\ \hline
      \makecell{R3 (use K3)} & I am a good journalist. \\ \hline
      \makecell{R4 (use K3)} & \makecell[l]{I also like reading and wish to be a jour-\\nalist, but now can only sell washers.} \\ \hline \hline
      Response & \makecell[l]{I love to write! Want to be journalist but\\ have settle for selling washers at sears.} \\ 
      \hline
\end{tabular}
\vspace{-0.2cm}
\caption{Comparison between Different Responses}
\vspace{-0.2cm}
\label{tab:cmp}
\end{table}

The problems caused by this discrepancy are illustrated in Table~\ref{tab:cmp},
which is a dialogue from \cite{zhang2018personalizing}.
In this dataset, each agent is associated with a persona profile, which is served as knowledge. 
Two agents exchange information based on the associated knowledge.
Given an utterance,
different responses can be generated depending on whether appropriate knowledge is used.
R1 utilizes no knowledge and thus ends up in a less informative response,
while other responses are more informative since they incorporate external knowledge.
However, among the knowledge,
both K1 and K3 are relevant to the utterance.
If we simply select knowledge based on the utterance (i.e, prior information) 
without knowing that K3 is used in the true response (i.e., posterior information),
it is difficult to generate a proper response since appropriate knowledge might not be selected.
If the model is trained by selecting wrong knowledge (e.g., K2 in R2) 
or knowledge irrelevant to the true response (e.g., K1), 
it can be seen that they are completely useless since they cannot provide any helpful information.
Note that it is also important to properly incorporate knowledge in response generation.
For example, though R3 selects correct knowledge K3, 
it results in a less relevant response due to inappropriate usage of knowledge.
Only R4 makes appropriate selection of knowledge and incorporates it properly in generating responses.

To tackle the aforementioned discrepancy,
we propose to separate the posterior distribution from the prior distribution. 
In the posterior distribution over knowledge, 
both utterances and response are utilized, 
while the prior distribution works without knowing responses in advance. 
Then, we try to minimize the distance between them.
Specifically,
during the training process,
our model is trained to minimize the KL divergence between the prior distribution and the posterior distribution
so that our model can approximate the posterior distribution accurately using the prior distribution. 
Then, during the inference process, 
the model samples knowledge merely based on the prior distribution
(i.e., without any posterior information)
and incorporates the sampled knowledge into response generation. 
It is proved that through this process,
the model can effectively learn to generate proper and informative responses by utilizing appropriate knowledge.

%
%
%
%
%

The contributions of this paper can be summarized below:
\begin{itemize}
    \item We clearly state and analyze the discrepancy between the prior and posterior distributions over knowledge in knowledge-grounded dialogue generation, 
    which has not been sufficiently studied in the previous work.
    \item We propose a novel neural model which separates the posterior distribution from the prior distribution.
    We prove that our knowledge selection mechanism is effective for appropriate response generation.
    \item Our comprehensive experiments demonstrate that our model significantly outperforms the existing ones by incorporating knowledge more properly and generating appropriate and informative responses.
\end{itemize}

\section{Model}
In this paper, we focus on training a neural model with an effective knowledge selection mechanism.
Given an utterance  $X = x_1x_2\ldots x_n$ ($x_t$ is the $t$-th word in $X$) 
and a collection of knowledge $\{K_i\}_{i=1}^N$ 
(where the ground-truth knowledge information is unknown),
the goal is to select appropriate knowledge from the collection and
to generate a response $Y = y_1 y_2\ldots y_m$ by incorporating the selected knowledge.

\begin{figure}[tb]
\centering
\includegraphics[width=8cm]{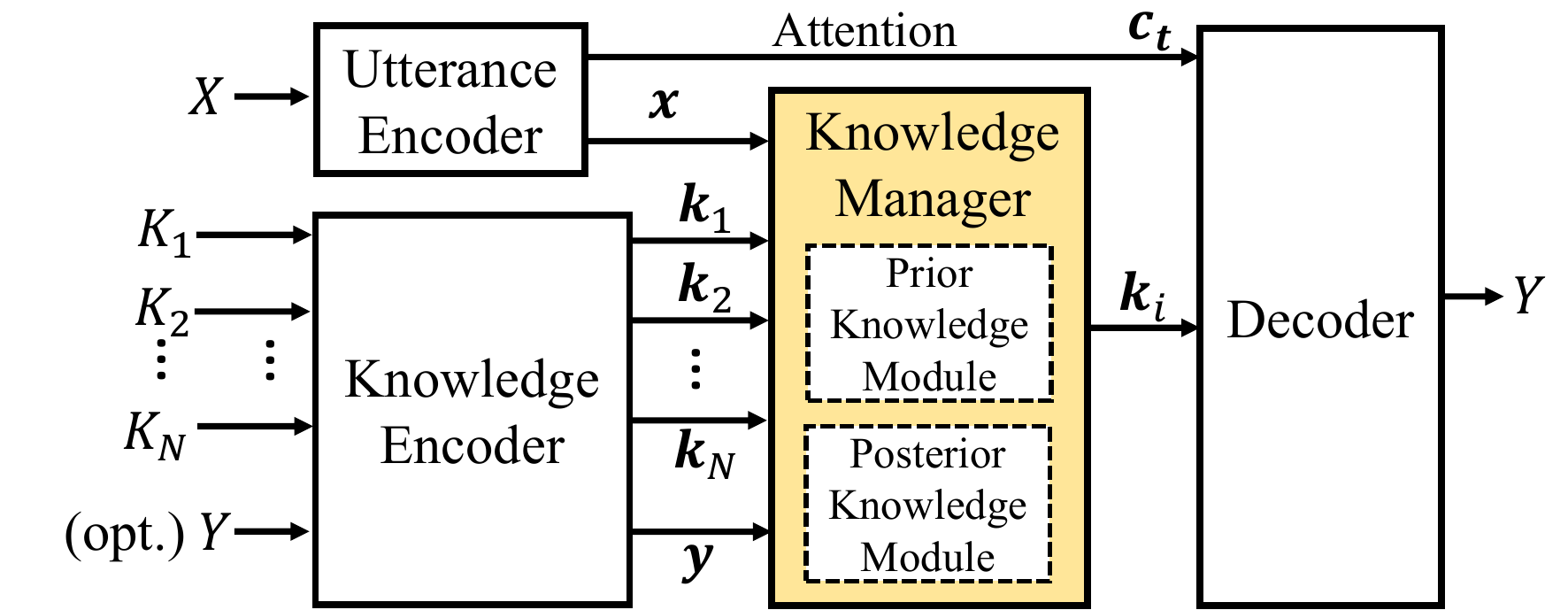}
\vspace{-0.2cm}
\caption{Architecture Overview}
\vspace{-0.2cm}
\label{fig:overview}
\end{figure}

\subsection{Architecture Overview}
The architecture overview of our model is presented in Figure~\ref{fig:overview}
and it consists of four major components:

\begin{itemize}
    \item \textbf{The utterance encoder} encodes $X$ into an utterance vector $\mathbf{x}$, 
    and feeds it into the knowledge manager.
    \item \textbf{The knowledge encoder} takes as input each knowledge $K_i$ and encodes it into a knowledge vector $\mathbf{k}_i$. 
    When response $Y$ is available, 
    it also encodes $Y$ into a vector~$\mathbf{y}$.
    \item \textbf{The knowledge manager} consists of two sub-modules: a prior knowledge module and a posterior knowledge module.
    Given the previously encoded $\mathbf{x}$ and $\{\mathbf{k}_i\}_{i=1}^N$ (and $\mathbf{y}$ if available),
    the knowledge manager is responsible to select an appropriate $\mathbf{k}_i$ and feeds it 
    (together with an attention-based context vector $\mathbf{c}_t$) into the decoder.
    \item \textbf{The decoder} generates responses based on the selected knowledge $\mathbf{k}_i$ and the attention-based context vector $\mathbf{c}_t$.
\end{itemize}

\subsection{Encoder}
\label{subsec:encoder}
We implement the utterance encoder 
using a bidirectional RNN with a gated recurrent unit (GRU) \cite{cho2014properties},
which consists of two parts: a forward RNN and a backward RNN.
Given utterance $X = x_1\ldots x_n$,
the forward RNN reads $X$ from left to right and then,
obtains a left-to-right hidden state $\overrightarrow{\mathbf{h}}_t$ for each $x_t$
while 
the backward RNN reads $X$ in a reverse order and similarly,
obtains a right-to-left hidden state $\overleftarrow{\mathbf{h}}_t$ for each $x_t$.
These two hidden states are concatenated to form an overall hidden state $\mathbf{h}_t$ for $x_t$:
$$\mathbf{h}_t = [\overrightarrow{\mathbf{h}}_t; \overleftarrow{\mathbf{h}}_t] = [\mathsf{GRU}(x_t, \overrightarrow{\mathbf{h}}_{t-1}); \mathsf{GRU}(x_t, \overleftarrow{\mathbf{h}}_{t+1})]$$
where
$[\cdot; \cdot]$ represents a vector concatenation. 
To obtain an encoded vector $\mathbf{x}$ for utterance $X$,
we utilize the hidden states and define $\mathbf{x} = [\overrightarrow{\mathbf{h}}_T ; \overleftarrow{\mathbf{h}}_1]$.
This vector will be fed into the knowledge manager to facilitate knowledge selection
and it will also serve as the initial hidden state of the decoder.

Our knowledge encoder follows the same structure as the utterance encoder, 
but they do not share any parameters.
Specifically,
it encodes each knowledge $K_i$ (and response $Y$ if available) 
into a vector $\mathbf{k}_i$ (and $\mathbf{y}$, respectively) using a bidirectional RNN
and uses it later in the knowledge manager.

\begin{figure}[tb]
\hspace{-0.5cm}
\includegraphics[width=9.3cm]{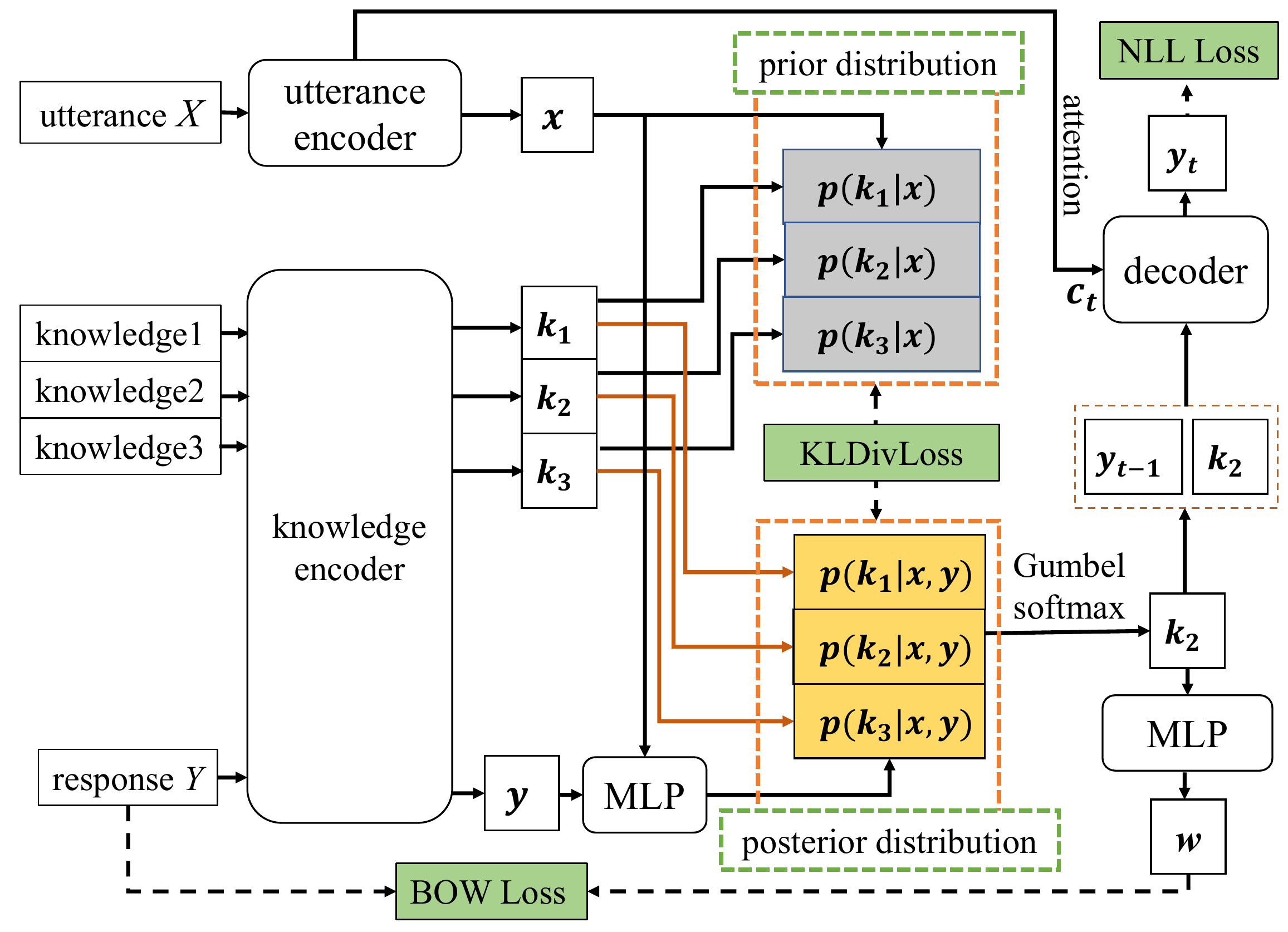}
\vspace{-0.7cm}
\caption{Knowledge Manager and Loss Functions}
\vspace{-0.2cm}
\label{fig:details}
\end{figure}

\subsection{Knowledge Manager}
\label{subsec:knowledge}
Given the encoded utterance $\mathbf{x}$ and the encoded knowledge collection $\{\mathbf{k}_i\}_{i=1}^N$,
the goal of the knowledge manager is to select an appropriate $\mathbf{k}_i$.
When the response $\mathbf{y}$ is available,
the model also utilized it to obtain $\mathbf{k}_i$. 
The knowledge manager consists of two sub-modules (see Figure~\ref{fig:details}): a prior knowledge module and a posterior knowledge module.

In the prior knowledge module,
we define a conditional probability distribution over knowledge,
denoted by $\mathbf{p}(\mathbf{k}|\mathbf{x})$:
$$p(\mathbf{k} = \mathbf{k}_i|\mathbf{x}) = \frac{\exp(\mathbf{k}_i \cdot \mathbf{x})}{\sum_{j=1}^N \exp(\mathbf{k}_j \cdot \mathbf{x})}$$
Intuitively, 
we use the dot product (i.e., attention \cite{bahdanau2014neural}) to measure the association between $\mathbf{k}_i$ and the input utterance $\mathbf{x}$.
A high association means that $\mathbf{k}_i$ is relevant to $\mathbf{x}$ and thus,
$\mathbf{k}_i$ is likely to be selected.
Note that $\mathbf{p}(\mathbf{k}|\mathbf{x})$ is conditioned \emph{only} on $\mathbf{x}$ and thus, 
it is a \emph{prior distribution over knowledge} since it works without knowing the response.
However,
there can be different knowledge that are relevant to the utterance,
and thus, it is difficult to select knowledge simply based on the prior distribution in training.

Motivated by this, in the posterior knowledge module, 
we define a \emph{posterior distribution over knowledge}, denoted by $\mathbf{p}(\mathbf{k}|\mathbf{x}, \mathbf{y})$, by considering both utterances and responses:
$$p(\mathbf{k} = \mathbf{k}_i|\mathbf{x}, \mathbf{y}) = \frac{ \exp(\mathbf{k}_i \cdot \mathsf{MLP}([\mathbf{x}; \mathbf{y}]))}{\sum_{j=1}^N \exp(\mathbf{k}_j \cdot \mathsf{MLP}([\mathbf{x}; \mathbf{y}]))}$$
where $\mathsf{MLP}(\cdot)$ is a fully connected layer.
Compared with the prior distribution, 
the posterior distribution is sharp
since the actual knowledge used in the true response $Y$ can be captured. 

According to the distributions defined above,
we sample knowledge 
using Gumbel-Softmax re-parametrization \cite{jang2016categorical}
(instead of the exact sampling)
since it allows back propagation in non-differentiable distributions.
Specifically, 
in the training process,
knowledge is sampled based on the posterior distribution,
which is inferred from the true response, and thus
it is more likely to obtain appropriate knowledge via this distribution.
In the inference process,
the posterior distribution is unknown since responses are not available.
Thus, knowledge is sampled based on the prior distribution.

Clearly, the discrepancy between prior and posterior distributions
introduces great challenges in training the model:
it is desirable to select knowledge based on the posterior distribution,
which, however, is unknown during inference.
In this paper,
we propose to approximate the posterior distribution using the prior distribution 
so that our model is capable to select appropriate knowledge even without posterior information.
For this purpose,
we introduce an auxiliary loss, namely the  Kullback-Leibler divergence loss (KLDivLoss), 
to measure the proximity between the prior distribution and the posterior distribution.
The KLDivLoss is defined as follows.

\smallskip
\noindent
\textbf{KLDivLoss.}
We define the KLDivLoss to be 
$$\mathcal{L}_{KL}(\theta)  = \sum_{i=1}^N p(\mathbf{k} = \mathbf{k}_i|\mathbf{x}, \mathbf{y}) \log \frac{p(\mathbf{k} = \mathbf{k}_i|\mathbf{x}, \mathbf{y})}{p(\mathbf{k} = \mathbf{k}_i|\mathbf{x})}$$
where $\theta$ denotes the model parameters.

When minimizing KLDivLoss,
the posterior distribution $\mathbf{p}(\mathbf{k}|\mathbf{x}, \mathbf{y})$ can be regarded as labels
and our model is instructed to use the prior distribution $\mathbf{p}(\mathbf{k}|\mathbf{x})$ to approximate $\mathbf{p}(\mathbf{k}|\mathbf{x}, \mathbf{y})$ accurately.
As a consequence,
even when the posterior distribution is unknown in the inference process
(since the actual response $Y$ is unknown),
the prior distribution $\mathbf{p}(\mathbf{k}|\mathbf{x})$ can be effectively utilized to sample appropriate knowledge so as to generate proper responses.
To the best of our knowledge, it is the first neural model, 
which incorporates the posterior distribution as guidance,
enabling accurate knowledge lookups and high quality response generation.

\subsection{Decoder}
\label{subsec:decoder}
The decoder generates response word by word sequentially 
by incorporating the selected knowledge $\mathbf{k}_i$.
We introduce two variants of decoders.
The first one is a ``hard'' decoder with a standard GRU
and the second one is ``soft'' decoder with a hierarchical gated fusion unit \cite{yao2017towards}.

\smallskip
\noindent
\textbf{Standard GRU with Concatenated Inputs.}
Let $\mathbf{s}_{t-1}$ be the last hidden state of the decoder, $y_{t-1}$ be the word generated in the last step
and $\mathbf{c}_t$ be an attention-based context vector of the encoder
(i.e., $\mathbf{c}_t = \sum_{i=1}^n \alpha_{t,i} \mathbf{h}_i$ where $\alpha_{t,i}$ measures the relevancy between $\mathbf{s}_{t-1}$ and the hidden state $\mathbf{h}_i$ of the encoder).
The hidden state of the decoder at time $t$ is:
$$
    \mathbf{s}_t = \mathsf{GRU}([y_{t-1}; \mathbf{k}_i], \mathbf{s}_{t-1}, \mathbf{c}_t)
$$
where we concatenate $y_{t-1}$ with $\mathbf{k}_i$.
This decoder is said to be a hard decoder since
$\mathbf{k}_i$ is forced to participate in decoding.

\smallskip
\noindent
\textbf{Hierarchical Gated Fusion Unit (HGFU).}
HGFU provides a softer way to incorporate knowledge into response generation and
it consists of three major components,
namely an utterance GRU, a knowledge GRU and a fusion unit.

The former two components follow the standard GRU structure, which
produce hidden representations for the last generated $y_{t-1}$ and the selected knowledge $\mathbf{k}_i$, respectively:
$$
    \mathbf{s}_t^y = \mathsf{GRU}(y_{t-1}, \mathbf{s}_{t-1}, \mathbf{c}_t)
\textrm{ and }
    \mathbf{s}_t^k = \mathsf{GRU}(\mathbf{k}_i, \mathbf{s}_{t-1}, \mathbf{c}_t)
$$
Then,
the fusion unit combines them to produce the hidden state of the decoder at time $t$ \cite{yao2017towards}:
$$ \mathbf{s}_t =  \mathbf{r} \odot \mathbf{s}_t^y + (\mathbf{1}- \mathbf{r}) \odot  \mathbf{s}_t^k$$
where
$\mathbf{r} = \sigma( \mathbf{W}_z[ \tanh(\mathbf{W}_y\mathbf{s}_t^y); \tanh(\mathbf{W}_k\mathbf{s}_t^k)])$ and $\mathbf{W}_z$, $\mathbf{W}_y$ and $\mathbf{W}_k$ are parameters.
Intuitively,
the gate $\mathbf{r}$ controls the contributions of $\mathbf{s}_t^y$ and $\mathbf{s}_t^k$ to the final hidden state $\mathbf{s}_t$,
allowing a flexible knowledge incorporation schema. 

\smallskip
After obtaining the hidden state $\mathbf{s}_t$,
the next word $y_t$ is generated according to the following probablity distribution:
$$y_t \sim \mathbf{p}_t = \mathsf{softmax}(\mathbf{s}_t, \mathbf{c}_t)$$

\subsection{Loss Function}
\label{subsec:loss}
Apart from the KLDivLoss,
two additional loss functions are used in our model:
the NLL loss captures the \emph{word order} information 
while the BOW loss captures the \emph{bag-of-word} information.
All loss functions are also elaborated in Figure~\ref{fig:details}.

\smallskip
\noindent
\textbf{NLL Loss.}
The objective of NLL loss is to quantify the difference between the true response and the response generated by our model.
It minimizes Negative Log-Likelihood (NLL):
$$\mathcal{L}_{NLL}(\theta) = -\mathbf{E}_{\mathbf{k}_i \sim \mathbf{p}(\mathbf{k} |\mathbf{x},\mathbf{y})}\sum_{t=1}^m  \log p(y_t| y_{<t}, \mathbf{x}, \mathbf{k}_i)$$
where $\theta$ denotes the model parameters and
$y_{<t}$ denotes the previously generated words.

\smallskip
\noindent
\textbf{BOW Loss.}
The BOW loss is adapted from  \cite{zhao2017learning} to ensure the accuracy of the sampled knowledge $\mathbf{k}_i$ 
by enforcing the relevancy between the knowledge and the true response.
Specifically, let $\mathbf{w} = \mathsf{MLP}(\mathbf{k_i})  \in \mathcal{R}^{|V|}$ where $|V|$ is the vocabulary size
and we define
$p(y_t | \mathbf{k}_i) = \frac{\exp(\mathbf{w}_{y_t})}{\sum_{v \in V}\exp(\mathbf{w}_{v})}$.
Then, the BOW loss is defined to minimize
$$\mathcal{L}_{BOW}(\theta) = -\mathbf{E}_{\mathbf{k}_i \sim \mathbf{p}(\mathbf{k} |\mathbf{x},\mathbf{y})}\sum_{t= 1}^m \log p(y_t | \mathbf{k}_i) $$

In summary, unless specified explicitly,
the total loss of a given a training example $(X, Y, \{K_i\}_{i=1}^N)$ is
$$\mathcal{L}(\theta) = \mathcal{L}_{KL}(\theta) + \mathcal{L}_{NLL}(\theta)+ \mathcal{L}_{BOW}(\theta)$$

\section{Experiments}
\subsection{Dataset}
We conducted experiments on two recently created datasets,
namely
the Persona-chat dataset \cite{zhang2018personalizing} and the Wizard-of-Wikipedia dataset \cite{dinan2018wizard}.

\smallskip
\noindent
In \textbf{Persona-chat},
each dialogue was constructed from a pair of  crowd-workers,
who chat to know each other.
To produce meaningful conversations,
each worker was assigned a persona profile, describing their characteristics,
and this profile serves as knowledge in the conversation.
There are 151,157 turns (each turn corresponds to an utterance and a response pair) of conversations in Persona-chat,
which we divide into 122,499 for train, 14,602 for validation and 14,056 for test.
The average size of a knowledge collection (the average number of sentences in a persona profile) in this dataset is 4.49.

\smallskip
\noindent
\textbf{Wizard-of-Wikipedia} is a chit-chatting dataset between two agents on some chosen topics.
One of the agent, also known as the wizard, plays the role of a knowledge expert and has access to a retrieval system for acquiring knowledge.
The other agent acts as a curious learner.
From this dataset, 79,925 turns of conversations are obtained 
and 68,931/3,686/7,308 of them are used for train/validation/test.
The test set is split into two subsets, Test Seen and Test Unseen.
Test Seen contains 3,619 turns of conversations on some overlapping topics with the training set,
while Test Unseen contains 3,689 turns on topics never seen before in train or validation.
Note that in this paper, we focus on the scenarios where ground-truth knowledge is unknown.
Thus, we did not use the ground-truth knowledge information provided in this dataset.
The average size of a knowledge collection accessed by the wizard is 67.57.

\begin{table*}[bt]
\centering
\small

\begin{tabular}{c|c|c|c|c|c} \hline
     \multirow{2}{*}{Dataset} &\multirow{2}{*}{Model} & \multicolumn{3}{c|}{Automatic Evaluation} & \multirow{2}{*}{ \makecell{Human\\Evaluation}} \\ \cline{3-5}
     & & BLEU-1/2/3 & Distinct-1/2 & Knowledge R/P/F1 & \\ \hline \hline
     \multirow{5}{*}{\makecell{Persona-chat} }
     & Seq2Seq & 0.182/0.093/0.055 & 0.026/0.074 & 0.0042/0.0172/0.0066 & 0.70 \\ \cline{2-6}
     & MemNet(hard) & 0.186/0.097/0.058 & 0.037/0.099 & 0.0115/0.0430/0.0175 & 0.79 \\ \cline{2-6}
     & MemNet(soft) & 0.177/0.091/0.055 & 0.035/0.096 & 0.0146/0.0567/0.0223 & 0.81 \\ \cline{2-6}
     & PostKS(concat) & 0.182/0.096/0.057 & \textbf{0.048}/0.126 & 0.0365/0.1486/0.0567 & 0.92 \\ \cline{2-6}
     & PostKS(fusion) & \textbf{0.190}/\textbf{0.098}/\textbf{0.059} & 0.046/\textbf{0.134} & \textbf{0.0574}/\textbf{0.2137}/\textbf{0.0870} & \textbf{0.97} \\ \hline \hline
     \multirow{5}{*}{\makecell{\makecell{Wizard-of-Wikipedia \\ (Test Seen)}}}
     & Seq2Seq & 0.169/0.066/0.032 & 0.036/0.112 & 0.0069/0.5780/0.0136 & 0.88 \\ \cline{2-6}
     & MemNet(hard) & 0.159/0.062/0.029 & 0.043/0.138 & 0.0077/0.6036/0.0151 & 0.93 \\ \cline{2-6}
     & MemNet(soft) & 0.168/0.067/0.034 & 0.037/0.115 & 0.0076/0.6713/0.0151 & 0.95 \\ \cline{2-6}
     & PostKS(concat) & 0.167/0.066/0.032 & \textbf{0.056}/0.209 & 0.0080/0.6979/0.0158 & 0.97 \\ \cline{2-6}
     & PostKS(fusion) & \textbf{0.172}/\textbf{0.069}/\textbf{0.034} & \textbf{0.056}/\textbf{0.213} & \textbf{0.0088}/\textbf{0.7047}/\textbf{0.0174} & \textbf{1.02}\\ \hline 
     \multirow{5}{*}{\makecell{\makecell{Wizard-of-Wikipedia \\ (Test Unseen)}}}
     & Seq2Seq & 0.150/0.054/0.026 & 0.020/0.063 & 0.0015/0.2052/0.0030 & 0.76 \\ \cline{2-6}
     & MemNet(hard) & 0.142/0.042/0.015 & 0.029/0.088 & 0.0025/0.3020/0.0050 & 0.79 \\ \cline{2-6}
     & MemNet(soft) & 0.148/0.048/0.023 & 0.026/0.081 & 0.0028/0.3793/0.0055 & 0.83 \\ \cline{2-6}
     & PostKS(concat) & 0.144/0.043/0.016 & \textbf{0.040}/0.151 & 0.0033/0.4392/0.0065 & 0.87 \\ \cline{2-6}
     & PostKS(fusion) & 0.147/0.046/0.021 & \textbf{0.040}/\textbf{0.156} & \textbf{0.0034}/\textbf{0.4772}/\textbf{0.0068} & \textbf{0.92} \\ \hline 
\end{tabular}
\vspace*{-0.2cm}
\caption{Automatic and Human Evaluation on Persona-chat and Wizard-of-Wikipedia}
\label{tab:evaluation}
\vspace*{-0.2cm}
\end{table*}

\begin{table*}[bt]
\centering
\small
\begin{tabular}{c|l|l} \cline{2-3}
      & Persona-chat & Wizard-of-Wikipedia \\ \hline \hline
      Utterance & \makecell[l]{I like all music. How about you?} & \makecell[l]{Cool! You sure know some stuff about country music!} \\ \hline \hline
      Knowledge  & 
      \makecell[l]{K1. I hate broccoli. \\ \hline
      K2. Rock music is my favorite. \\ \hline
      K3. I am afraid of the dark.
      } & 
      \makecell[l]{
      K1. George Glenn Jones (September 12, 1931 – April 26, 2013) was\\ an American musician, singer and songwriter. \\ \hline
      K2. In 2009, in the United States, country music was the most listened\\ to rush hour radio genre.\\ \hline
      K3. Country (or country and western) is a musical genre that originated\\ in the southern United States in the early 1920s. 
      }\\ \hline \hline
      Seq2Seq & I am good. How are you? & Yes, I know country music. \\ \hline
      \makecell{MemNet(hard)} & \makecell[l]{I don't like broccoli. What about you?} & I love rock music. It was one of the most popular rock bands in UK.  \\ \hline
      \makecell{MemNet(soft)} & \makecell[l]{I like all kinds of music. What do you do? } & I do know that country music was originated in the United States. \\ \hline 
     \makecell{PostKS(concat)} & I like to listen to rock music. & I love George Glenn. He was an American singer and songwriter. \\ \hline
     \makecell{PostKS(fusion)} & \makecell{I love rock music. What is your favorite band?} & I like country music. It is the most listened to rush hour radio genre.\\ \hline
\end{tabular}
\vspace*{-0.2cm}
\caption{Examples of the Generated Responses on Persona-chat and Wizard-of-Wikipedia}
\vspace*{-0.2cm}
\label{tab:case}
\end{table*}

\begin{table*}[bt]
\centering
\small
\begin{tabular}{c|c|c|c|c|c|c} \hline
     \multirow{2}{*}{Dataset} &\multirow{2}{*}{Model} & \multicolumn{4}{c|}{Automatic Evaluation} & \multirow{2}{*}{ \makecell{Human\\Evaluation}} \\ \cline{3-6}
     & & PPL & BLEU-1/2/3 & Distinct-1/2 & Knowledge R/P/F1 & \\ \hline \hline
     \multirow{2}{*}{\makecell{Persona-chat} }
     & LIC & 31.4 & 0.169/0.072/0.035 & 0.112/0.435 & 0.0308/0.0956/0.0446 & 1.26 \\ \cline{2-7}
     & LIC+PostKS & \textbf{30.5} & \textbf{0.180}/\textbf{0.081}/\textbf{0.040} & \textbf{0.118}/\textbf{0.470} & \textbf{0.1043}/\textbf{0.3423}/\textbf{0.1529} & \textbf{1.33} \\ \hline \hline
     \multirow{2}{*}{\makecell{\makecell{Wizard-of-Wikipedia}\\ (Test Seen) }}
     & LIC & 64.7 & 0.161/0.065/0.032 & 0.119/0.491 & 0.0151/0.7308/0.0297 & 1.18 \\ \cline{2-7}
     & LIC+PostKS & \textbf{59.8} & \textbf{0.167/0.068/0.034} & \textbf{0.121/0.502} & \textbf{0.0233/0.7676/0.0452} & \textbf{1.30} \\ \hline
     \multirow{2}{*}{\makecell{\makecell{Wizard-of-Wikipedia}\\ (Test Unseen) }}
     & LIC & 96.8 & 0.144/0.042/0.015 & 0.105/0.411 & 0.0124/0.6832/0.0244 & 0.97  \\ \cline{2-7}
     & LIC+PostKS & \textbf{91.8} & \textbf{0.148/0.046/0.02} & \textbf{0.113/0.442} & \textbf{0.0147/0.7109/0.0289} & \textbf{1.12} \\ \hline
\end{tabular}
\vspace*{-0.2cm}
\caption{Lost in Conversation with our Knowledge Selection Mechanism}
\label{tab:lic}
\vspace*{-0.2cm}
\end{table*}

\subsection{Models for Comparison}
We implemented our model, namely the \emph{\underline{Post}erior \underline{K}nowledge \underline{S}election (PostKS)} model, for evaluation. 
In particular, two variants of our model were implemented to demonstrate the effect of different ways of incorporating knowledge:
\begin{itemize}
    \item \textbf{PostKS(concat)}: the hard knowledge-grounded model with a GRU decoder where knowledge is concatenated.
    \item \textbf{PostKS(fusion)}: the soft knowledge-grounded model where knowledge is incorporated with a HGFU.
\end{itemize}

We compared our models with three baselines:
\begin{itemize}
    \item \textbf{Seq2Seq}: an attention Seq2Seq that does not have access to external knowledge  \cite{vinyals2015neural}. 
    \item \textbf{MemNet(hard)}: a memory network from \cite{ghazvininejad2018knowledge}, where knowledge is sampled based on prior semantic similarity and fed into the decoder.
    \item \textbf{MemNet(soft)}: a soft knowledge-grounded model from \cite{ghazvininejad2018knowledge},
    where knowledge is stored in memory units that are decoded with attention.
\end{itemize}
In our adaption of all baselines, we used the same RNN encoder/decoder as PostKS.
Among them, Seq2Seq is compared for demonstrating the effect of introducing knowledge in response generation while
MemNet based models, which also have access to knowledge, are compared to verify that the effectiveness of our novel knowledge selection mechanism.

\subsection{Implementation Details}
Our encoders and decoders have 2-layer GRU structures with 800 hidden states for each layer,
but they do not share any parameters.
We set 
the word embedding size to be 300 and 
initialized it using GloVe \cite{pennington2014glove}.
The vocabulary size is 20,000.
We used the Adam optimizer with a mini-batch size of 128 and
the learning rate is 0.0005.

We trained our model with at most 20 epochs on a P40 machine.
In the first 5 epochs,
we minimize the BOW loss only for pre-training the knowledge manager.
In the remaining epochs,
we minimize over the sum of all losses.
After each epoch, we save a model and 
the model with the minimum loss is selected for evaluation.
Our models and datasets are all available online: \url{https://github.com/ifr2/PostKS}.

\subsection{Automatic and Human Evaluation}
We adopted several automatic metrics to perform evaluation
and the result is summarized in Table~\ref{tab:evaluation}.
Among them,
\emph{BLEU-1/2/3} and \emph{Distinct-1/2} are two widely used metrics for evaluating the quality and diversity of generated responses.
\emph{Knowledge R/P/F1} is a metric adapted from \cite{dinan2018wizard},
which measures the unigram recall/precision/F1 score between the generated responses and the knowledge collection.
Specifically,
given the set of non-stopwords in $Y$ and in the knowledge collection $\{K_i\}_{i=1}^N$, denoted by 
$W_Y$ and $W_K$,
we define Knowledge R(ecall) and Knowledge P(recision) to be
$$|W_Y\cap W_K|/|W_K| \textrm{ and } |W_Y\cap W_K|/|W_Y|$$
and Knowledge F1 = $2 \cdot \frac{\textrm{Recall}\cdot \textrm{Precision}}{\textrm{Recall} + \textrm{Precision}}$.

As shown in Table~\ref{tab:evaluation},
our models outperform all baselines \emph{significantly} (p $<0.00001$)
by achieving the \emph{highest} scores in most of the automatic metrics. 
Specifically,
compared with Seq2Seq,
incorporating knowledge is shown to be helpful in generating diverse responses.
For example, Distinct-1/2 on Persona-chat is increased
from 0.026/0.074 (Seq2Seq) to 0.048/0.126 (PostKS(concat)),
meaning that the diversity is greatly improved by augmenting with knowledge.
Besides, when comparing with existing knowledge-grounded baselines,
our models demonstrate their ability on incorporating appropriate knowledge in response generation.
In particular, comparing PostKS(fusion) against MemNet(soft) on Persona-chat 
(they are soft knowledge-grounded models except that we use both prior and posterior information to facilitate knowledge selection),
we achieve higher BLEU and Distinct scores. 
This is because that the posterior information is better utilized in our models to provide effective guidance on obtaining appropriate knowledge,
resulting in responses with better quality. 
Compared with knowledge selection on Persona-chat,
selecting appropriate knowledge on Wizard-of-Wikipedia is more challenging  
due to a larger knowledge collection size.
Nevertheless,
our models perform consistently better than most baselines.
For example, PostKS(fusion) has higher knowledge R/P/F1 compared with all MemNet based models on Wizard-of-Wikipedia,
indicating that it can not only select appropriate knowledge, 
but also ensure that knowledge is better incorporated in the response generated.
Finally,
we observe that PostKS(fusion) performs slightly better than PostKS(concat) in most cases.
This verifies that soft knowledge incorporation is a better way of introducing knowledge to response generation 
since it allows for more flexible knowledge integration and less sensitivity to noise.

In human evaluation,
three annotators were recruited to rate the overall quality of the responses generated by each model. 
The rating ranges from 0 to 2,
where 0 means that the response is completely irrelevant,
1 means that the response is acceptable but not very informative, and
2 means that the response is natural, relevant and informative.
We randomly sampled 300 responses for each model on each dataset,
resulting in 4,500 responses in total for human annotation.
We reported the average rating in Table~\ref{tab:evaluation}. 
The agreement ratio (Fleiss' kappa \cite{fleiss1971measuring})
is 0.48 and 0.41 on Persona-chat and Wizard-of-Wikipedia,
showing moderate agreement.
According to the result, 
both of our models, PostKS(concat) and PostKS(fusion), are remarkably better than all existing baselines in terms of human rating,
demonstrating the effectiveness of our novel knowledge selection mechanism. 

\subsection{Case Study}
Table~\ref{tab:case} shows two example responses.
For the lack of space, 
we only display three pieces of knowledge on each dataset.
In the example from Persona-chat,
the utterance is asking whether the agent likes \emph{music}.
Without access to external knowledge,
Seq2Seq produces a bland response which does not contain any useful information.
MemNet(hard) tries to incorporate knowledge,
but, unfortunately, it selects the wrong knowledge, 
leading to an irrelevant response about \emph{broccoli} rather than \emph{music}.
The remaining models generate responses with the help of the correct knowledge.
Among them, our PostKS(fusion) and PostKS(concat) models perform better
since they are more specific by mentioning exactly the \emph{rock music}.
In particular, our soft knowledge-grounded model, PostKS(fusion), performs noticeably well
since it does not only answer questions, but also raises a relevant question about the \emph{favorite band}, allowing evolving conversations. 
The example from Wizard-of-Wikipedia is about \emph{country music}.
Similar to Persona-chat,
our models enjoy superior performance by producing informative and relevant responses. 

\subsection{Further Evaluation of Knowledge Selection}
To further verify the effectiveness our knowledge selection mechanism,
we apply it on the best performing Transformer model, \underline{L}ost \underline{i}n \underline{C}onversation (LIC), in ConvAI2 NeurIPS competition \cite{dinan2019second}
and the result is reported in Table~\ref{tab:lic} (following \cite{dinan2018wizard}, Perplexity (PPL) is reported). 
After integrating our mechanism,
all metrics are greatly improved.
In particular, we achieve a threefold improvement on knowledge R/P/F1 on Persona-chat,
which verifies the usefulness of our knowledge selection mechanism in incorporating knowledge in response generation.

\section{Related Work}
The success of Seq2Seq motivates the development of various techniques for improving the quality of generated responses.
Examples include diversity promotion \cite{li2016diversity} and unknown words handling \cite{gu2016incorporating}.
However, the problem of tending to generate generic words still remains 
since they do not have the access to external information.

Recently,
knowledge incorporation is shown to be an effective way to improve the performance of neural models.
%
Long et al.~\shortcite{long2017knowledge} obtained knowledge from texts using a convolutional network.
Ghazvininejad et al.~\shortcite{ghazvininejad2018knowledge} stored texts as knowledge in a memory network to produce more informative responses.
A knowledge diffusion model was also proposed in \cite{liu2018knowledge},
where the model is augmented with divergent thinking over a knowledge base.
Large scale commonsense knowledge bases were first utilized in \cite{zhou2018commonsense} and
many domain-specific knowledge bases were also considered to ground neural models with knowledge \cite{xu2017incorporating,zhu2017flexible,gu2016incorporating}. 

However, most existing knowledge-grounded models condition knowledge simply on conversation history,
which we regard as a prior distribution over knowledge.
Compared with the posterior distribution over knowledge, 
which further considers the actual knowledge used in the true responses,
the prior distribution has a larger variance
and thus, existing models can hardly select appropriate knowledge simply based on the prior distribution during the training process.
In comparison, we carefully analyze the discrepancy between prior and posterior distributions
and
our model has been effectively taught 
to select appropriate knowledge and to ensure that knowledge is better utilized in generating responses.

Our work is related to conditional variation autoencoders (CVAE) \cite{zhao2017learning}
where a \emph{recognition network} is used to approximate a posterior distribution,
but we have the following differences.
Firstly, we focus on different problems.
In this paper, we focus on knowledge-grounded conversations where our model employs a novel knowledge selection mechanism
while CVAE aims at capturing the \emph{discourse-level diversity}. 
Secondly,
CVAE learns a distribution in a \emph{latent space}
where the meanings of latent variables are difficult to interpret,
while we \emph{explicitly} define the distributions over knowledge based on the semantic similarity on utterances and responses,
which has better understandability.

\section{Conclusion}
In this paper, we present a model with a novel knowledge selection mechanism,
which is the first neural model that makes use of both prior and posterior distributions over knowledge to facilitate knowledge selection.
We analyze the discrepancy between prior and posterior distributions,
which has not been studied before.
By effectively approximating the posterior distribution using the prior distribution,
our model can generate appropriate responses during inference.
Extensive experiments on both automatic and human metrics demonstrate the effectiveness and usefulness of our model.
As for future work,
we plan to extend our knowledge selection mechanism for selecting knowledge in multi-turn conversations.

\section*{Acknowledgments}
We would like to thank Siqi Bao, Chaotao Chen and Huang He
for their help and valuable suggestions on this paper.

\bibliography{ref}

\begin{thebibliography}{}

\bibitem[\protect\citeauthoryear{Bahdanau \bgroup \em et al.\egroup
  }{2014}]{bahdanau2014neural}
Dzmitry Bahdanau, Kyunghyun Cho, and Yoshua Bengio.
\newblock Neural machine translation by jointly learning to align and
  translate.
\newblock {\em arXiv preprint arXiv:1409.0473}, 2014.

\bibitem[\protect\citeauthoryear{Cho \bgroup \em et al.\egroup
  }{2014a}]{cho2014properties}
Kyunghyun Cho, Bart van Merrienboer, Dzmitry Bahdanau, and Yoshua Bengio.
\newblock On the properties of neural machine translation: Encoder-decoder
  approaches.
\newblock In {\em Eighth Workshop on Syntax, Semantics and Structure in
  Statistical Translation}, pages 103--111, 2014.

\bibitem[\protect\citeauthoryear{Cho \bgroup \em et al.\egroup
  }{2014b}]{cho2014learning}
Kyunghyun Cho, Bart van Merrienboer, {\c{C}}aglar G{\"{u}}l{\c{c}}ehre, Dzmitry
  Bahdanau, Fethi Bougares, Holger Schwenk, and Yoshua Bengio.
\newblock Learning phrase representations using {RNN} encoder-decoder for
  statistical machine translation.
\newblock In {\em Proceedings of the 2014 Conference on Empirical Methods in
  Natural Language Processing}, pages 1724--1734, 2014.

\bibitem[\protect\citeauthoryear{Dinan \bgroup \em et al.\egroup
  }{2018}]{dinan2018wizard}
Emily Dinan, Stephen Roller, Kurt Shuster, Angela Fan, Michael Auli, and Jason
  Weston.
\newblock Wizard of wikipedia: Knowledge-powered conversational agents.
\newblock {\em arXiv preprint arXiv:1811.01241}, 2018.

\bibitem[\protect\citeauthoryear{Dinan \bgroup \em et al.\egroup
  }{2019}]{dinan2019second}
Emily Dinan, Varvara Logacheva, Valentin Malykh, Alexander Miller, Kurt
  Shuster, Jack Urbanek, Douwe Kiela, Arthur Szlam, Iulian Serban, Ryan Lowe,
  et~al.
\newblock The second conversational intelligence challenge.
\newblock {\em arXiv preprint arXiv:1902.00098}, 2019.

\bibitem[\protect\citeauthoryear{Fleiss}{1971}]{fleiss1971measuring}
Joseph~L Fleiss.
\newblock Measuring nominal scale agreement among many raters.
\newblock {\em Psychological bulletin}, page 378, 1971.

\bibitem[\protect\citeauthoryear{Ghazvininejad \bgroup \em et al.\egroup
  }{2018}]{ghazvininejad2018knowledge}
Marjan Ghazvininejad, Chris Brockett, Ming-Wei Chang, Bill Dolan, Jianfeng Gao,
  Wen-tau Yih, and Michel Galley.
\newblock A knowledge-grounded neural conversation model.
\newblock In {\em Thirty-Second AAAI Conference on Artificial Intelligence},
  2018.

\bibitem[\protect\citeauthoryear{Gu \bgroup \em et al.\egroup
  }{2016}]{gu2016incorporating}
Jiatao Gu, Zhengdong Lu, Hang Li, and Victor O.~K. Li.
\newblock Incorporating copying mechanism in sequence-to-sequence learning.
\newblock In {\em Proceedings of the 54th Annual Meeting of the Association for
  Computational Linguistics}, 2016.

\bibitem[\protect\citeauthoryear{Jang \bgroup \em et al.\egroup
  }{2016}]{jang2016categorical}
Eric Jang, Shixiang Gu, and Ben Poole.
\newblock Categorical reparameterization with gumbel-softmax.
\newblock {\em arXiv preprint arXiv:1611.01144}, 2016.

\bibitem[\protect\citeauthoryear{Li \bgroup \em et al.\egroup
  }{2016}]{li2016diversity}
Jiwei Li, Michel Galley, Chris Brockett, Jianfeng Gao, and Bill Dolan.
\newblock A diversity-promoting objective function for neural conversation
  models.
\newblock In {\em The 2016 Conference of the North American Chapter of the
  Association for Computational Linguistics}, pages 110--119, 2016.

\bibitem[\protect\citeauthoryear{Liu \bgroup \em et al.\egroup
  }{2018}]{liu2018knowledge}
Shuman Liu, Hongshen Chen, Zhaochun Ren, Yang Feng, Qun Liu, and Dawei Yin.
\newblock Knowledge diffusion for neural dialogue generation.
\newblock In {\em Proceedings of the 56th Annual Meeting of the Association for
  Computational Linguistics}, pages 1489--1498, 2018.

\bibitem[\protect\citeauthoryear{Long \bgroup \em et al.\egroup
  }{2017}]{long2017knowledge}
Yinong Long, Jianan Wang, Zhen Xu, Zongsheng Wang, Baoxun Wang, and Zhuoran
  Wang.
\newblock A knowledge enhanced generative conversational service agent.
\newblock In {\em Proceedings of the 6th Dialog System Technology Challenges
  (DSTC6) Workshop}, 2017.

\bibitem[\protect\citeauthoryear{Pennington \bgroup \em et al.\egroup
  }{2014}]{pennington2014glove}
Jeffrey Pennington, Richard Socher, and Christopher Manning.
\newblock Glove: Global vectors for word representation.
\newblock In {\em Proceedings of the 2014 conference on empirical methods in
  natural language processing}, pages 1532--1543, 2014.

\bibitem[\protect\citeauthoryear{Shang \bgroup \em et al.\egroup
  }{2015}]{shang2015neural}
Lifeng Shang, Zhengdong Lu, and Hang Li.
\newblock Neural responding machine for short-text conversation.
\newblock In {\em Proceedings of the 53rd Annual Meeting of the Association for
  Computational Linguistics and the 7th International Joint Conference on
  Natural Language Processing of the Asian Federation of Natural Language
  Processing}, pages 1577--1586, 2015.

\bibitem[\protect\citeauthoryear{Vinyals and Le}{2015}]{vinyals2015neural}
Oriol Vinyals and Quoc Le.
\newblock A neural conversational model.
\newblock {\em arXiv preprint arXiv:1506.05869}, 2015.

\bibitem[\protect\citeauthoryear{Xu \bgroup \em et al.\egroup
  }{2017}]{xu2017incorporating}
Zhen Xu, Bingquan Liu, Baoxun Wang, Chengjie Sun, and Xiaolong Wang.
\newblock Incorporating loose-structured knowledge into conversation modeling
  via recall-gate {LSTM}.
\newblock In {\em 2017 International Joint Conference on Neural Networks},
  pages 3506--3513, 2017.

\bibitem[\protect\citeauthoryear{Yao \bgroup \em et al.\egroup
  }{2017}]{yao2017towards}
Lili Yao, Yaoyuan Zhang, Yansong Feng, Dongyan Zhao, and Rui Yan.
\newblock Towards implicit content-introducing for generative short-text
  conversation systems.
\newblock In {\em Proceedings of the 2017 Conference on Empirical Methods in
  Natural Language Processing}, pages 2190--2199, 2017.

\bibitem[\protect\citeauthoryear{Zhang \bgroup \em et al.\egroup
  }{2018}]{zhang2018personalizing}
Saizheng Zhang, Emily Dinan, Jack Urbanek, Arthur Szlam, Douwe Kiela, and Jason
  Weston.
\newblock Personalizing dialogue agents: {I} have a dog, do you have pets too?
\newblock In {\em Proceedings of the 56th Annual Meeting of the Association for
  Computational Linguistics}, pages 2204--2213, 2018.

\bibitem[\protect\citeauthoryear{Zhao \bgroup \em et al.\egroup
  }{2017}]{zhao2017learning}
Tiancheng Zhao, Ran Zhao, and Maxine Esk{\'{e}}nazi.
\newblock Learning discourse-level diversity for neural dialog models using
  conditional variational autoencoders.
\newblock In {\em Proceedings of the 55th Annual Meeting of the Association for
  Computational Linguistics}, pages 654--664, 2017.

\bibitem[\protect\citeauthoryear{Zhou \bgroup \em et al.\egroup
  }{2018}]{zhou2018commonsense}
Hao Zhou, Tom Young, Minlie Huang, Haizhou Zhao, Jingfang Xu, and Xiaoyan Zhu.
\newblock Commonsense knowledge aware conversation generation with graph
  attention.
\newblock In {\em Proceedings of the Twenty-Seventh International Joint
  Conference on Artificial Intelligence}, pages 4623--4629, 2018.

\bibitem[\protect\citeauthoryear{Zhu \bgroup \em et al.\egroup
  }{2017}]{zhu2017flexible}
Wenya Zhu, Kaixiang Mo, Yu~Zhang, Zhangbin Zhu, Xuezheng Peng, and Qiang Yang.
\newblock Flexible end-to-end dialogue system for knowledge grounded
  conversation.
\newblock {\em arXiv preprint arXiv:1709.04264}, 2017.

\end{thebibliography}
\bibliographystyle{named}

\end{document}